\documentclass[letterpaper]{article}
\usepackage{aaai}
\usepackage{times}
\usepackage{helvet}
\usepackage{courier}

\usepackage{amsmath}
\usepackage{mathrsfs}
\usepackage[round]{natbib}
\usepackage{graphicx}
\begin{document}
\title{MEMEN: Multi-layer Embedding with Memory Networks for Machine Comprehension}

\author{Boyuan Pan$^*_{\bigtriangleup}$, Hao Li$^*_{\bigtriangleup}$, Zhou Zhao$^\dag$, Bin Cao$^\ddagger$, Deng Cai$^*$, Xiaofei He$^*$\\
	$^*$State Key Lab of CAD$\&$CG, College of Computer Science, Zhejiang University, Hangzhou, China\\
	$^\dag$College of Computer Science, Zhejiang University, Hangzhou, China\\
	$^\ddagger$Eigen Technologies, Hangzhou, China\\
	$\{$panby, haolics, zhaozhou$\}$@zju.edu.cn, bincao@aidigger.com, $\{$dengcai, xiaofeihe$\}$@gmail.com,  \\
}

\maketitle

\begin{abstract}
\begin{quote}
Machine comprehension(MC) style question answering is a representative problem in natural language processing. Previous methods rarely spend time on the improvement of encoding layer, especially the embedding of syntactic information and name entity of the words, which are very crucial to the quality of encoding. Moreover, existing attention methods represent each query word as a vector or use a single vector to represent the whole query sentence, neither of them can handle the proper weight of the key words in query sentence. In this paper, we introduce a novel neural network architecture called Multi-layer Embedding with Memory Network(MEMEN) for machine reading task. In the encoding layer, we employ classic skip-gram model to the syntactic and semantic information of the words to train a new kind of embedding layer. We also propose a memory network of full-orientation matching of the query and passage to catch more pivotal information. Experiments show that our model has competitive results both from the perspectives of precision and efficiency in Stanford Question Answering Dataset(SQuAD) among all published results and achieves the state-of-the-art results on TriviaQA dataset.

\end{quote}
\end{abstract}

\section{Introduction}
Machine comprehension(MC) has gained significant popularity over the past few years and it is a coveted goal in the field of natural language processing and artificial intelligence~\citep{shen2016reasonet,seo2016bidirectional}. Its task is to teach machine to understand the content of a given passage and then answer the question related to it. Figure 1 shows a simple example from the popular dataset SQuAD \citep{rajpurkar2016squad}.

\begin{figure}[htbp]

  \includegraphics[width=0.48 \textwidth]{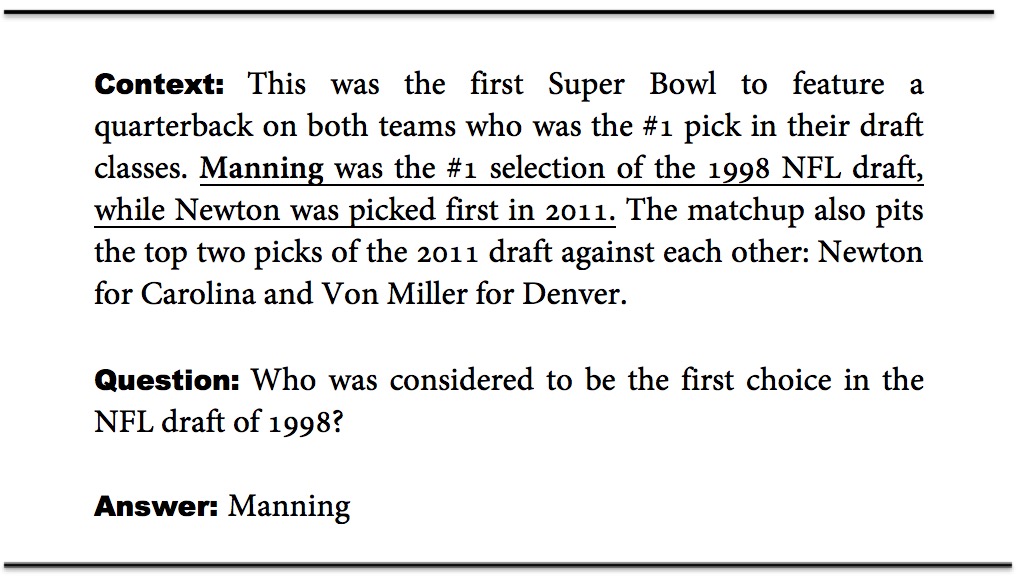}
\caption{An example from the SQuAD dataset. The answer is a segment text of context.}
\end{figure}

Many significant works are based on this task, and most of them focus on the improvement of a sequence model that is augmented with an attention mechanism. However, the encoding of the words is also crucial and a better encoding layer can lead to substantial difference to the final performance. Many powerful methods~\citep{hu2017mnemonic,rnet,seo2016bidirectional} only represent their words in two ways, word-level embeddings and character-level embeddings. They use pre-train vectors, like \emph{GloVe}\citep{pennington2014glove}, to do the word-level embeddings, which ignore syntactic information and name entity of the words. \citet{liu2017structural} construct a sequence of syntactic nodes for the words and encodes the sequence into a vector representation. However, they neglected the optimization of the initial embedding and didn't take the semantic information of the words into account, which are very important parts in the vector representations of the words. For example, the word ``Apple" is a fixed vector in \emph{GloVe} and noun in syntactics whatever it represents the fruit or the company, but name entity tags can help recognize.

Moreover, the attention mechanism can be divided into two categories: one dimensional attention~\citep{chen2016thorough,dhingra2016gated,kadlec2016text} and two dimensional attention~\citep{cui2016attention,wang2016multi,seo2016bidirectional}. In one dimensional attention, the whole query is represented by one embedding vector, which is usually the last hidden state in the neural network. However, using only one vector to represent the whole query will attenuate the attention of key words. On the contrary, every word in the query has its own embedding vector in the situation of two dimensional attention, but many words in the question sentence are useless even if disturbing, such as the stopwords.

\begin{figure*}[htbp]
  \begin{center}
  \includegraphics[width=1 \textwidth]{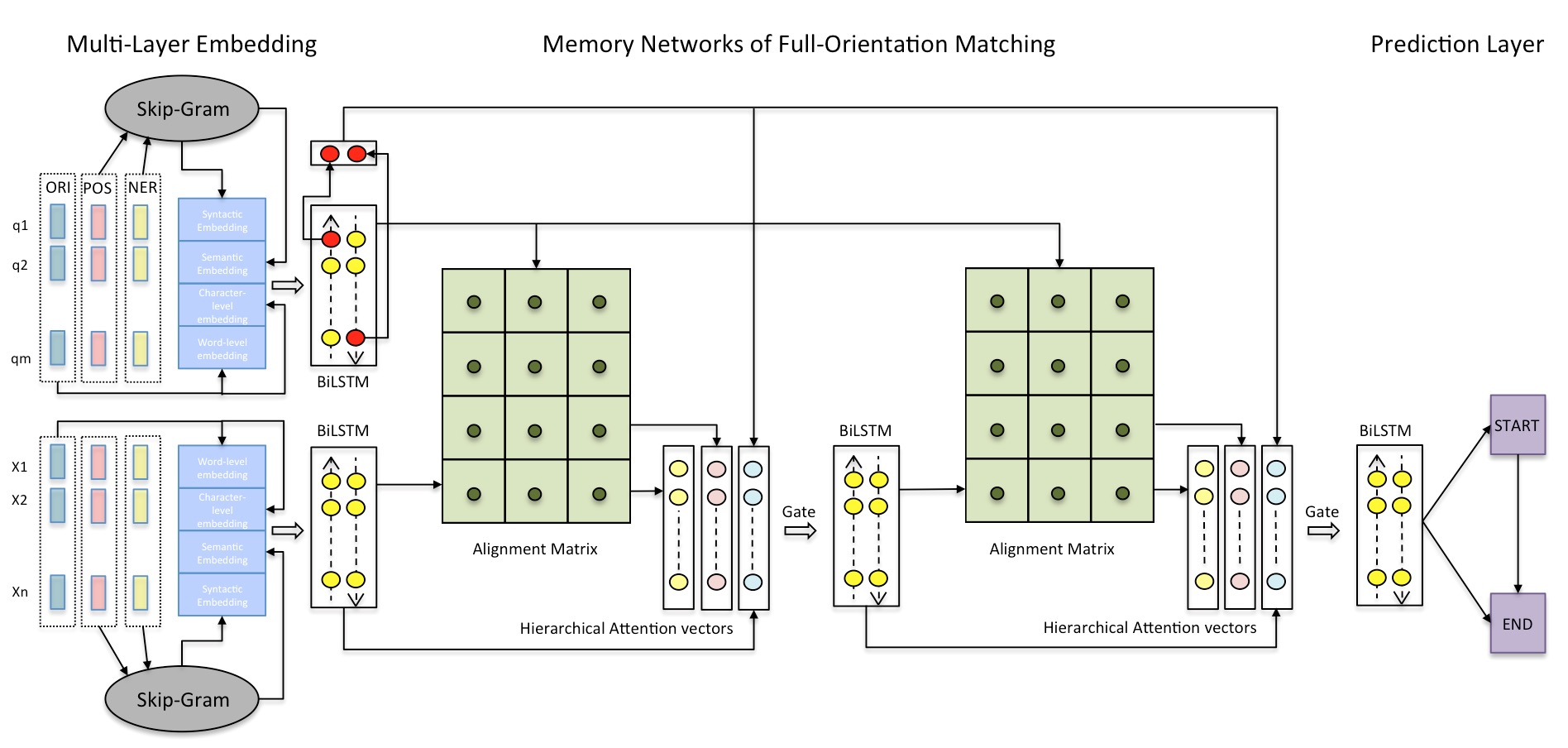}
  \caption{MEMEN structure overview. In the figure above, there are two hops in the memory network.}
  \end{center}
\end{figure*}

In this paper, we introduce the Multi-layer Embedding with Memory Networks(MEMEN), an end-to-end neural network for machine comprehension task. Our model consists of three parts: 1) the encoding of context and query, in which we add useful syntactic  and semantic information in the embedding of every word, 2) the high-efficiency multi-layer memory network of full-orientation matching to match the question and context, 3) the pointer-network based answer boundary prediction layer to get the location of the answer in the passage. The contributions of this paper can be summarized as follows.

\begin{itemize}
\item First, we propose a novel multi-layer embedding of the words in the passage and query. We use skip-gram model to train the part-of-speech(POS) tags and name-entity recognition(NER) tags embedding that represent the syntactic and semantic information of the words respectively. The analogy inference provided by  skip-gram model can make the similar attributes close in their embedding space such that more adept at helping find the answer.

\item Second, we introduce a memory networks of full-orientation matching.To combines the advantages of one dimensional attention and two dimensional attention, our novel hierarchical attention vectors contain both of them. Because key words in query often appear at ends of the sentence, one-dimensional attention, in which the bi-directional last hidden states are regarded as representation, is able to capture more useful information compared to only applying two dimensional attention. In order to deepen the memory and  better understand the passage according to the query, we employ the structure of multi-hops to repeatedly read the passage. Moreover, we add a gate to the end of each memory to improve the speed of convergence.

\item Finally, the proposed method yields competitive results on the large machine comprehension bench marks SQuAD and the state-of-the-art results on TriviaQA dataset. On SQuAD, our model achieves 75.37\% exact match and 82.66\% F1 score. Moreover, our model avoids the high computation complexity self-matching mechanism which is popular in many previous works, thus we spend much less time and memory when training the model. 

\end{itemize}

\begin{figure*}[htbp]
  \begin{center}
  \includegraphics[width=0.8 \textwidth]{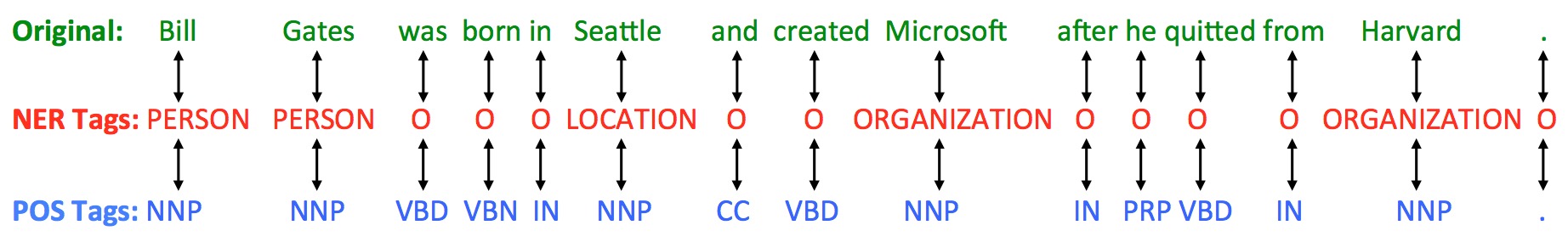}
  \caption{The passage and its according transformed ``passages". The first row(green) is the original sentence from the passage, the second row(red) is the name-entity recognition(NER) tags, and the last row(blue) is the part-of-speech(POS) tags.}
  \end{center}
\end{figure*}

\section{Model Structure}
As Figure 2 shows, our machine reading model consists of three parts. First, we concatenate several layers of embedding of questions and contexts and pass them into a bi-directional RNN\citep{mikolov2010recurrent}. Then we obtain the relationship between query and context through a novel full-orientation matching and apply memory networks in order to deeply understand. In the end, the output layer helps locate the answer in the passage.
\subsection{Encoding of Context and Query}
In the encoding layer, we represent all tokens in the context and question as a sequence of embeddings and pass them as the input to a recurrent neural network. 

Word-level embeddings and character-level embeddings are first applied.We use pre-trained word vectors \emph{GloVe}\citep{pennington2014glove} to obtain the fixed word embedding of each word.The character-level embeddings are generated by using Convolutional Neural Networks(CNN) which is applied to the characters of each word(Kim, et al., 2014). This layer maps each token to a high dimensional vector space and is proved to be helpful in handling out-of-vocab(OOV) words.

We also use skip-gram model to train the embeddings of part-of-speech(POS) tags and named-entity recognition(NER) tags. We first transform all of the given training set into their part-of-speech(POS) tags and named-entity recognition(NER) tags, which can be showed in Figure 3. Then we employ skip-sram model, which is one of the core algorithms in the popular off-the-shelf embedding \emph{word2vec}\citep{mikolov2013distributed}, to the transformed ``passage'' just like it works in \emph{word2vec} for the normal passage. Given a sequence of training words in the transformed passage: $w_1, w_2, ..., w_N$, the objective of the skip-gram model is to maximize the average log probability:
$$\frac{1}{N} \sum_{n = 1}^{N} \sum_{-c \leq i \leq c, p \neq 0} log~p(w_{n+i}|w_n)$$
where $c$ is the size of the context which can be set manually, a large $c$ means more accurate results and more training time. The $p(w_{n+i}|w_n)$ is defined by:
$$p(w_O|w_I) = \frac{exp({v'}_{w_O}^{T} v_{w_I})}{\sum_{w = 1}^{V} exp({v'}_{w}^{T} v_{w_I})}$$
where $v_w$ and ${v'}_{w}$ are the input and output vector of $w$, and $V$ is the vocabulary size. 

We finally get the fixed length embedding of each tag. Although the number of tags limits the effect of word analogy inference, it still be very helpful compared to simple one hot embedding since similar tags have similar surroundings.

In the end, we use a BiLSTM to encode both the context and query embeddings and obtain their representations $\{ r^{P}_{t} \}^{n}_{t=1}$ and $\{ r^{Q}_{t} \}^{m}_{t=1}$ and the last hidden state of both directions of query representation $u^{Q}$.
 $$r^{P}_{t} = {\rm BiLSTM}([w^{P}_{t}; c^{P}_{t}; s^{P}_{t}]), t \in [1,...,n]$$
 $$r^{Q}_{t} = {\rm BiLSTM}([w^{Q}_{t}; c^{Q}_{t}; s^{Q}_{t}]), t \in [1,...,m]$$
where $w$, $c$, $s$ represent word-level embedding, character-level embedding and tags embedding respectively. $u^{Q}$ is the concatenation of both directions' last hidden state.

\subsection{Memory Network of Full-Orientation Matching}
Attention mechanism is a common way to link and blend the content between the context and query. Unlike previous methods that are either two dimensional matching or one dimensional matching, we propose a full-orientation matching layer that synthesizes both of them and thus combine the advantages of both side and hedge the weakness. After concatenating all the attention vectors, we will pass them into a bi-directional LSTM. We start by describing our model in the single layer case, which implements a single memory hop operation. We then show it can be stacked to give multiple hops in memory. \\

\subsubsection{Integral Query Matching}
The input of this step is the representations $\{ r^{P}_{t} \}^{n}_{t=1}$, $\{ r^{Q}_{t} \}^{m}_{t=1}$ and $u^{Q}$. At first we obtain the importance of each word in passage according to the integral query by means of computing the match between $u^{Q}$ and each representation  $ r^{P}_{t}$  by taking the inner product followed by a softmax:
$$c_t = {\rm softmax} (\langle u^{Q}, r^{P}_{t} \rangle)$$

Subsequently the first matching module is the sum of the inputs $\{ r^{P}_{t} \}^{n}_{t=1}$ weighted by attention $c_t$:
$$m^{1} = \sum_t c_t r^{P}_{t}$$

\subsubsection{Query-Based Similarity Matching}
We then obtain an alignment matrix $A \in R^{n \times m}$ between the query and context by $A_{ij} = w_{1}^{T}[r^{P}_{i};r^{Q}_{j};r^{P}_{i} \circ r^{Q}_{j}]$, $w_1$ is the weight parameter, $\circ$ is elementwise multiplication. Like Seo et al. (2017), we use this alignment matrix to compute whether the query words are relevant to each context word. For each context word, there is an attention weight that represents how much it is relevant to every query word:
$$B = {\rm softmax}_{row} (A) \in R^{n \times m}$$

${\rm softmax}_{row}(A)$ means the softmax function is performed across the row vector, and each attention vector is $M^{2}_t= B \cdot r^{Q}_{t} $, which is based on the query embedding. Hence the second matching module is $M^2$, where each $M^{2}_t$ is the column of $M^2$.\\

\subsubsection{Context-Based Similarity Matching}
When we consider the relevance between context and query, the most representative word in the query sentence can be chosen by $e = {\rm max}_{row}(A) \in R^{n} $, and the attention is $d = {\rm softmax}(e)$. Then we obtain the last matching module
$$m^{3} = \sum_{t} r^{P}_{t}  \cdot d_t$$

which is based on the context embedding. We put all of the memories in a linear function to get the integrated hierarchical matching module:
$$M = f(M^1, M^2, M^3)$$

where $f$ is an simple linear function, $M^1$ and $M^3$ are matrixes that are tiled n times by $m^1$ and $m^3$. \\

Moreover, \cite{rnet} add an additional gate to the input of RNN:
$$g_t = {\rm sigmoid} (W_g M)$$
$$M^{*} = g_t \odot M$$

The gate is based on the integration of hierarchical attention vectors, and it effectively filtrates the part of tokens that are helpful in understanding the relation between passage and query. Additionally, we add a bias to improve the estimation:
$$g_t = {\rm sigmoid} (W_g M + b)$$

Experiments prove that this gate can also accelerate the speed of convergence. Finally, the integrated memory $M$ is passed into a bi-directional LSTM, and the output will captures the interaction among the context words and the query words:
$$O_t = {\rm BiLSTM} (O_{t-1}, M)$$

In multiple layers, the integrated hierarchical matching module $M$ can be regarded as the input $\{ r^{P}_{t} \}^{n}_{t=1}$ of next layer after a dimensionality reduction processing. We call this memory networks of full-orientation matching.

\subsection{Output layer}
In this layer, we follow \citet{wang2016machine} to use the boundary model of pointer networks \citep{vinyals2015pointer} to locate the answer in the passage. Moreover, we follow \cite{rnet} to initialize the hidden state of the pointer network by a query-aware representation:
$$z_j = s^T {\rm tanh} (W^{Q} r^{Q}_{j} + b^{Q})$$
$$a_i = \frac{{\rm exp}(z_i)}{\sum^{m}_{j = 1} {\rm exp}(z_j)}$$
$$l^{0} = \sum^{m}_{i = 1}a_i r^{Q}_{i}$$
where $ s^T$,$W^{Q}$ and $b^{Q}$ are parameters, $l^{0}$ is the initial hidden state of the pointer network. Then we use the passage representation along with the initialized hidden state to predict the indices that represent the answer's location in the passage:
$$z^{k}_{j} = c^T {\rm tanh} (W^{P} O_{j} + W^{h} l^{0})$$
$$a^{k}_{i} = \frac{{\rm exp}(z^{k}_i)}{\sum^{n}_{j = 1} {\rm exp}(z^{k}_j)}$$
$$p^{k} = {\rm argmax}(a^{k}_{1}, ..., a^{k}_{n})$$
where $W^{h}$ is parameter, $k = 1,2$ that respectively represent the start point and the end point of the answer, $O_{j}$ is the vector that represents $j$-th word in the passage of the final output of the memory networks. To get the next layer of hidden state, we need to pass $O$ weighted by current predicted probability $a^{k}$ to the Gated Recurrent Unit(GRU)(Chung et al., 2014):
$$v^{k} = \sum_{i = 1}^{n} a^{k}_{i}O_{i}$$
$$l^{1}_{t} = {\rm GRU}(l^{1}_{t-1}, v^{k})$$
For the loss function, we minimize the sum of the negative probabilities of the true start and end indices by the predicted distributions.

\section{Experiment}
\subsection{Implementation Settings}
The tokenizers we use in the step of preprocessing data are from Stanford CoreNLP (Manning et al., 2014). We also use part-of-speech tagger and named-entity recognition tagger in Stanford CoreNLP utilities to transform the passage and question. For the skip-gram model, our model refers to the \emph{word2vec} module in open source software library, \emph{Tensorflow}, the skip window is set as 2. The dataset we use to train the embedding of POS tags and NER tags are the training set given by SQuAD, in which all the sentences are tokenized and regrouped as a list. To improve the reliability and stabllity, we screen out the sentences whose length are shorter than 9. We use 100 one dimensional filters for CNN in the character level embedding, with width of 5 for each one. We set the hidden size as 100 for all the LSTM and GRU layers and apply dropout\citep{srivastava2014dropout} between layers with a dropout ratio as 0.2. We use the AdaDelta \citep{zeiler2012adadelta} optimizer with a initial learning rate as 0.001. For the memory networks, we set the number of layer as 3.

 \subsection{TriviaQA Results}
We first evaluate our model on a large scale reading comprehension dataset TriviaQA version1.0\citep{joshi2017triviaqa}. TriviaQA contains over 650K question-answer-evidence triples, that are derived from Web search results and Wikipedia pages – with highly differing levels of information redundancy. TriviaQA is the first dataset where questions are authored by trivia enthusiasts, independently of the evidence documents. 

\begin{figure}[htbp]
  \begin{center}
  \includegraphics[width=0.4 \textwidth]{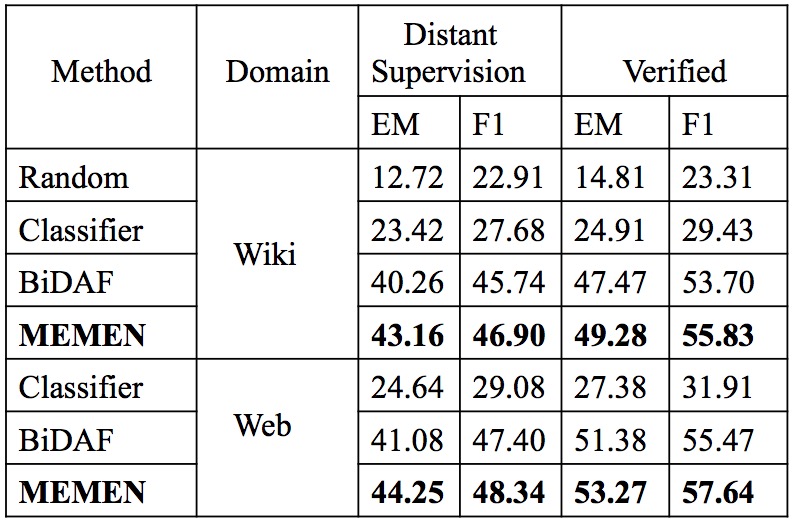}
  \caption{The performance of our MEMEN and baselines on TriviaQA dataset.}
  \end{center}
\end{figure}

  \begin{figure*}[htbp]
  \begin{center}
  \includegraphics[width=0.7 \textwidth]{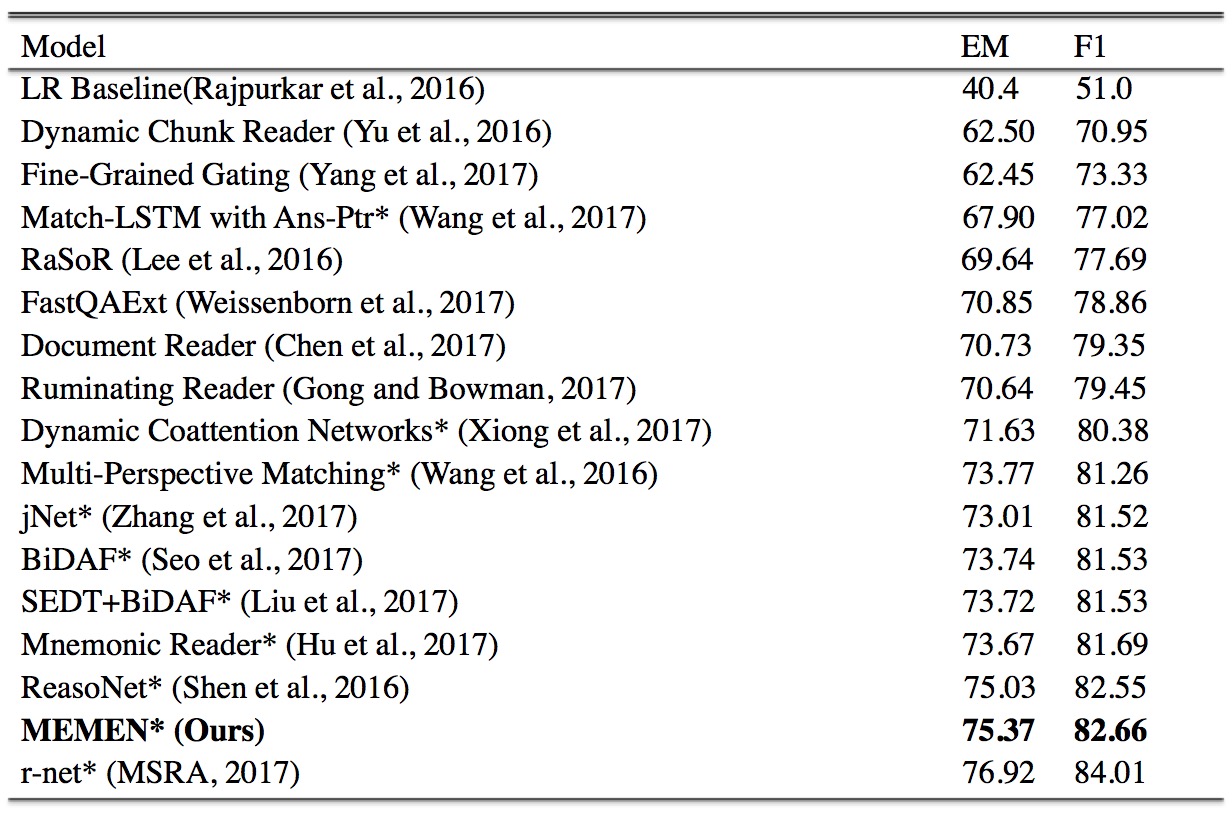}
  \caption{The performance of our MEMEN and competing approaches on SQuAD dataset as we submitted our model (May, 22, 2017). * indicates ensemble models.}
  \end{center}
\end{figure*}

There are two different metrics to evaluate model accuracy: Exact Match(EM) and F1 Score, which measures the weighted average of the precision and recall rate at character level. Because the evidence is gathered by an automated process, the documents are not guaranteed to contain all facts needed to answer the question. In addition to distant supervision evaluation, we also evaluate models on a verified subsets. Because the test set is not released, we train our model on training set and evaluate our model on dev set. As we can see in Figure 4, our model outperforms all other baselines and achieves the state-of-the-art result on all subsets on TriviaQA.

\subsection{SQuAD Results}
We also use the Stanford Question Answering Dataset (SQuAD) v1.1 to conduct our experiments. Passages in the dataset are retrieved from English Wikipedia by means of Project Nayuki's Wikipedia's internal PageRanks. They sampled 536 articles uniformly at random with a wide range of topics, from musical celebrities to abstract concepts.The dataset is partitioned randomly into a training set(80\%), a development set(10\%), and a hidden test set(10\%). The host of SQuAD didn't release the test set to the public, so everybody has to submit their model and the host will run it on the test set for them. 

Figure 5 shows the performance of our model and competing approaches on the SQuAD. The results of this dataset are all exhibited on a leaderboard, and top methods are almost all ensemble models, our model achieves an exact match score of 75.37\% and an F1 score of 82.66\%, which is competitive to state-of-the-art method. 

\subsection{Ensemble Details}
The main current ensemble methods in the machine comprehension is simply choosing the answer with the highest sum of confidence scores among several single models which are exactly identical except the random initial seed. However, the performance of ensemble model can obviously be better if there is some diversity among single models. In our SQuAD experiment, we get the value of learning rate and dropout ratio of each model by a gaussian distribution, in which the mean value are 0.001 and 0.2 respectively. 
$$learning~rate \sim N(0.001, 0.0001) $$
$$dropout \sim N(0.2, 0.05)$$
To keep the diversity in a reasonable scope, we set the variance of gaussian distribution as $0.0001$ and $0.05$ respectively. Finally, we build an ensemble model which consists of 14 single models with different parameters.

\subsection{Speed and Efficiency}
Compared to \citep{rnet}, which achieves state-of-the-art result on the SQuAD test set, our model doesn't contain the self-matching attention layer which is stuck with high computational complexity. Our MEMEN was trained with NVIDIA Titan X GPU, and the training process of the 3-hops model took roughly 5 hours on a single GPU. However, an one-hop model took 22 hours when we added self-matching layer in attention memory. Although the accuracy is improved a little compared to one-hop MEMEN model, it declined sharply as the number of hops increased, not to speak of the disadvantage of running time. The reason might be that multi-hops model with self-matching layer is too complex to efficiently learn the features for this dataset. As a result, our model is competitive both in accuracy and efficiency.

\subsection{Hops and Ablations}
Figure 6 shows the performance of our single model on SQuAD dev set with different number of hops in the memory network. As we can see, both the EM and F1 score increase as the number of hops enlarges until it arrives 3. After the model achieves the best performance with 3 hops, the performance gets worse as the number of hops gets large, which might result in overfitting. 

We also run the ablations of our single model on SQuAD dev set to evaluate the individual contribution. As Figure 7 shows, both syntactic embeddings and semantic embeddings contribute towards the model's performance and the POS tags seem to be more important. The reason may be that the number of POS tags is larger than that of NER tags so the embedding is easier to train. For the full-orientation matching, we remove each kind of attention vector respectively and the linear function can handle any two of the rest hierarchical attention vectors. For ablating integral query matching, the result drops about 2\% on both metrics and it shows that the integral information of query for each word in passage is crucial. The query-based similarity matching accounts for about 10\% performance degradation, which proves the effectiveness of alignment context words against query. For context-based similarity matching, we simply took out the 
$M^3$ from the linear function and it is proved to be contributory to the performance of full-orientation matching.

\begin{figure}[htbp]
  \begin{center}
  \includegraphics[width=0.24 \textwidth]{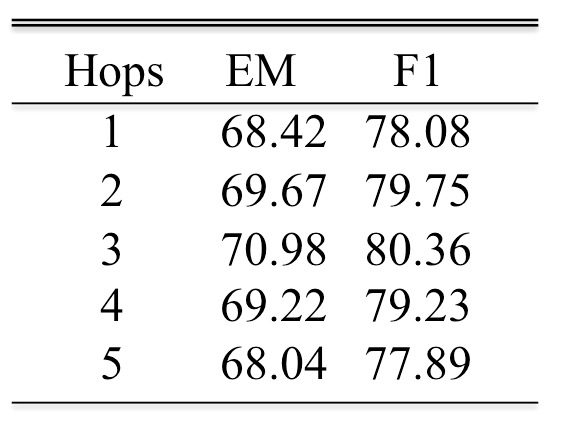}
  \caption{Performance comparison among different number of hops on the SQuAD dev set.}
  \end{center}
\end{figure}

\begin{figure}[htbp]
  \begin{center}
  \includegraphics[width=0.5 \textwidth]{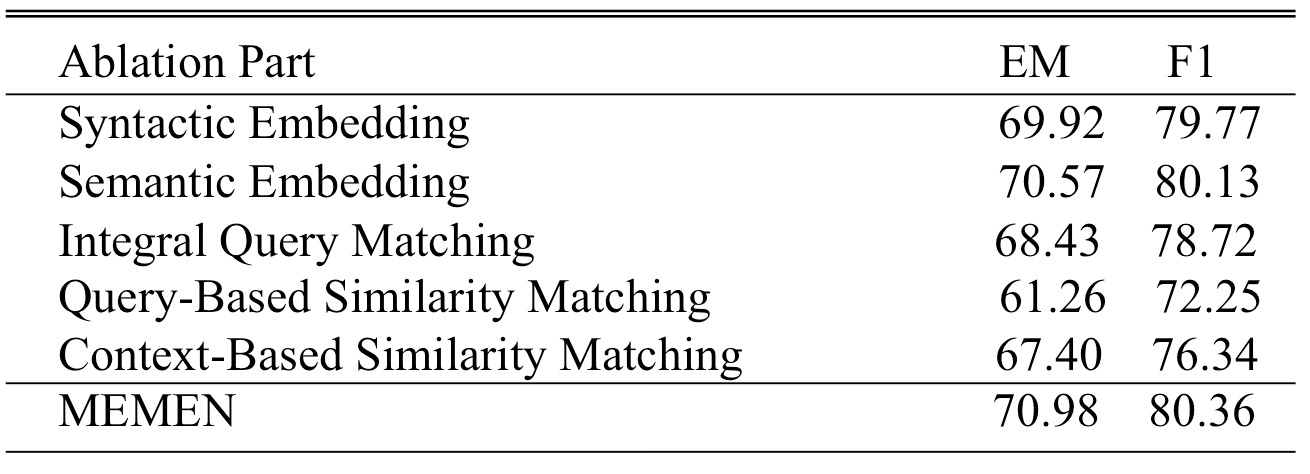}
  \caption{Ablation results on the SQuAD dev set.}
  \end{center}
\end{figure}

\section{Related Work}
\subsection{Machine Reading Comprehension Dataset. }
Several benchmark datasets play an important role in progress of machine comprehension task and question answering research in recent years. MCTest\citep{richardson2013mctest} is one of the famous and high quality datasets. There are 660 fictional stories and 4 multiple choice questions per story contained in it, and the labels are all made by humans. Researchers also released cloze-style datasets\citep{hill2015goldilocks,hermann2015teaching,onishi2016did,paperno2016lambada}. However, these datasets are either not large enough to support deep neural network models or too easy to challenge natural language.

Recently, \citet{rajpurkar2016squad} released the Stanford Question Answering dataset (SQuAD), which is almost two orders of magnitude larger than all previous hand-annotated datasets. Moreover, this dataset consists 100,000+ questions posed by crowdworkers on a set of Wikipedia articles, where the answer to each question is a segment of text from the corresponding passage, rather than a limited set of multiple choices or entities. TriviaQA \citep{joshi2017triviaqa} is also a large and high quality dataset, and the crucial difference between TriviaQA and SQuAD is that TriviaQA questions have not been crowdsourced from pre-selected passages.

\subsection{Attention Based Models for Machine Reading}
Many works are based on the task of machine reading comprehension, and attention mechanism have been particularly successful\citep{xiong2016dynamic,cui2016attention,wang2016multi,seo2016bidirectional,hu2017mnemonic,shen2016reasonet,rnet}. \citet{xiong2016dynamic} present a coattention encoder and dynamic decoder to locate the answer. \citet{cui2016attention} propose a two side attention mechanism to compute the matching between the passage and query. \citet{wang2016multi} match the passage and query from several perspectives and predict the answer by globally normalizing probability distributions. \citet{seo2016bidirectional} propose a bi-directional attention flow to achieve a query-aware context representation.  \citet{hu2017mnemonic} propose self-aware representation and multi-hop query-sensitive pointer to predict the answer span. \citet{shen2016reasonet} propose iterarively inferring the answer with a dynamic number of steps trained with reinforcement learning. \citet{rnet} employ gated self-matching attention to obtain the relation between the question and passage. Our MEMEN construct a hierarchical orientation attention mechanism to get a wider match while applying memory network\citep{sukhbaatar2015end} for deeper understand.

\section{Conclusion}
In this paper, we introduce MEMEN for Machine comprehension style question answering. We propose the multi-layer embedding to encode the document and the memory network of full-orientation matching to obtain the interaction of the context and query.  The experimental evaluation shows that our model achieves the state-of-the-art result on TriviaQA dataset and competitive result in SQuAD. Moreover, the ablations and hops analysis demonstrate the importance of every part of the hierarchical attention vectors and the benefit of multi-hops in memory network. For future work, we will focus on question generative method and sentence ranking in machine reading tasks.

\bibliographystyle{named}
\bibliography{memen_aaai}

\end{document}